\documentclass[letterpaper, 10 pt, conference]{ieeeconf}  

\IEEEoverridecommandlockouts                              

\usepackage{graphics} 
\usepackage{epsfig} 
\usepackage{mathptmx} 
\usepackage{times} 
\usepackage{amsmath} 
\usepackage{amssymb}  
\usepackage[caption=false]{subfig}
\usepackage{multirow}

\title{\LARGE \bf
Multimodal Interaction with Multiple Co-located Drones \\ in Search and Rescue Missions
}

\author{Jonathan Cacace$^1$,  
Alberto Finzi$^1$, Vincenzo Lippiello$^{1}$
\thanks{$^1$ Universit\`a degli Studi di Napoli Federico II, 
       via Claudio 21, 80125, Naples, Italy.
       {\tt\small \{jonathan.cacace, alberto.finzi, lippiell\}@unina.it}}
}

\begin{document}

\maketitle
\thispagestyle{empty}
\pagestyle{empty}

\begin{abstract}

We present a multimodal interaction framework suitable for a human rescuer that operates in proximity with a set of co-located drones during search  missions. This work is framed in the context of the SHERPA project whose goal is to develop a mixed ground and aerial robotic platform to support search and rescue activities in a real-world alpine scenario. Differently from typical human-drone interaction settings, here the operator is not fully dedicated to the drones, but involved in search and rescue tasks, hence only able to provide sparse, incomplete, although high-value, instructions to the robots. This operative scenario requires a human-interaction framework that supports multimodal communication along with an effective and natural mixed-initiative interaction between the human and the robots. In this work, we illustrate the domain and the proposed multimodal interaction framework discussing the system at work in a simulated case study.
\end{abstract}

\section{INTRODUCTION}

A multimodal interaction framework suitable for human-UAVs interaction in search and rescue missions is presented in this paper. 
This work is framed in  the context of the SHERPA project \cite{Sherpa2} whose goal is to develop a mixed ground and aerial robotic platform supporting search and rescue (SAR) activities in a real-world alpine scenario. 
One of the peculiar aspects of the SHERPA domain is the presence of a special rescue operator, called the {\em busy genius}, that should cooperate with a team of aerial vehicles in order to accomplish the rescue tasks.  
%
%
Differently from typical human-UAVs interaction scenarios \cite{Cummings2007,Bitton2008,MazaCMPO10}, in the place of a fully dedicated operator we have a rescuer which might be deeply involved in the SAR mission, hence only able to provide fast, incomplete, sparse, although high-value, inputs to the robots. In this context, the human should focus his cognitive effort on relevant and critical activities (e.g. visual inspection, precise maneuvering, etc.), 
while relying on the robotic autonomous system for specialized tasks (navigation, scan, etc.). Moreover, the human should operate in proximity with the drones in hazardous scenarios (e.g. avalanche), hence the required interaction 
is substantially dissimilar to the one considered in other works where the human and co-located UAVs cooperate in controlled indoor conditions \cite{PfeilKL13,szafir2014communication,NagGiuGamDic14:iros,MonajjemiWVM13}. 
%
%
%
In this paper, we present the multimodal and mixed-initiative interaction framework we are currently designing for this challenging and novel domain. The multimodal interaction should allow the operator to communicate with the robots in a natural, incomplete, but robust manner exploiting gestures, vocal, or tablet-based commands. 
Currently, we are manly focusing on a gesture- and speech-based interaction suitable for accomplishing navigation and search tasks in coordination with a set of drones operating in the scene.  In order to communicate with the robots, we assume the human equipped with light and wearable devices, such as a headset (speech) and the \textit{Myo Gesture Control Armband}\footnote{https://www.thalmic.com/en/myo/} (gestures).
Notice that, vision-based interaction/recognition methods, like the one proposed in \cite{PfeilKL13,MonajjemiWVM13,NagGiuGamDic14:iros}, are not appropriate in our context. 
In this domain, we introduce a set of multimodal commands and communication primitives suitable for the accomplishment of cooperative search tasks. In the proposed framework both {\em command-based} and {\em joystick-based} interaction metaphors can be exploited and smoothly combined to affect the robots behavior. 
%
The proposed framework permits different kinds of interactions, from precise vocal commands (e.g. the operator can ask the robot to  ``{\em rotate 3 o'clock}'' or  ``{\em go up 3 meters}''), to deictic communication where speech and gestures are combined (e.g. the operator can say  ``{\em go there}'' while pointing). Moreover, while a robot is executing a task, the human can exploit gestures in {\em joystick-based} metaphor to adjust the generated trajectory or to directly teleoperate the selected drones. Indeed, the interpreted human interventions are continuously integrated within the robotic control loop by a mixed-initiative system that can adjust the drone behaviors according to the operator intentions \cite{iros14,BevacquaCFL15}.
In order to test the framework and the associated interaction modalities, we introduce a test-bed where a human operator can orchestrate the operations of simulated drones to search for lost people in an alpine scenario. 
This case study is used to discuss the functioning and the effectiveness of the proposed interaction system.

\section{SEARCH AND RESCUE MISSIONS WITH MULTIPLE DRONES}
\begin{figure}[h]
	\begin{center}
		\includegraphics[scale=0.26]{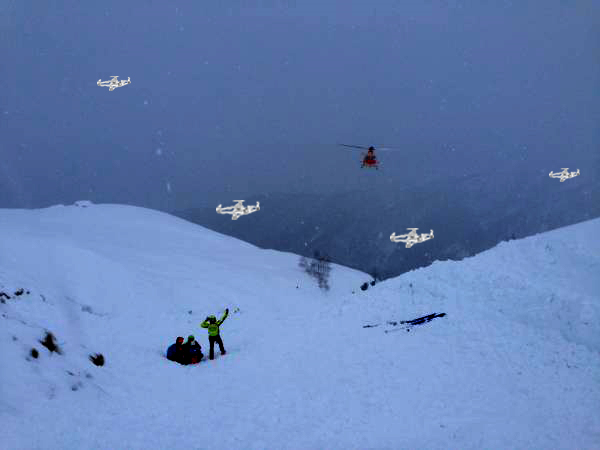}
		\par
		\caption[Scenario]{An illustration of the SHERPA winter scenario.}
		\label{fig:scenario}
	\end{center}
\end{figure}

We assume a human operator involved in SAR tasks in an alpine environment with the support of a set of co-located drones (see Fig. \ref{fig:scenario}). The mission goal is to find a set of missing persons in a specified area with loose time constraints. In particular, we refer to a winter scenario where the probability of survival decreases dramatically with the time. Moreover, we focus on the search phase of the rescue mission where the rescuer and the drones are already in the operative area. During the search, the operator can issue verbal and/or gestural commands to the drones which are used to extend the rescuers’ perception by streaming video or images taken with their on-board cameras. We assume a restricted search area of few square kilometers ($1$ to $10$) with a short mission time (less than $15$ minutes). As for the drones, we assume a set of small quadrotors with standard specification (flight time 25 min., max. airspeed 15 m/s, max. climb rate 8 m/s, etc.) equipped with standard sensors including an onboard camera used by the operator to remotely inspect the environment. In order to communicate with the drones, we assume the following (light and low-cost) human equipment: a tablet with a user interface, a headset to vocally communicate, and a {\em Thalmic Myo Armband} device, endowed with Eight Steel EMG and 9 $DOF$ IMU, for gesture-based interaction and teleoperation.

\section{MIXED-INITIATIVE MULTIMODAL INTERACTION}
In the domain illustrated above, the operator should interact with the robots in a rapid, concise, and natural manner, exploiting verbal and non-verbal communication. 
Moreover, since the human is not fully dedicated to the drones, the robotic system should support different control modes sliding form an {\em autonomous} behavior to direct {\em teleoperation}, passing through the {\em mixed-initiative} mode, where the user can execute some operations while relying on the autonomous control system for the remaining ones. 

\subsection{Multimodal Mixed-Initiative Interaction Architecture}

\begin{figure}[h]
	\begin{center}
		\includegraphics[scale=0.4]{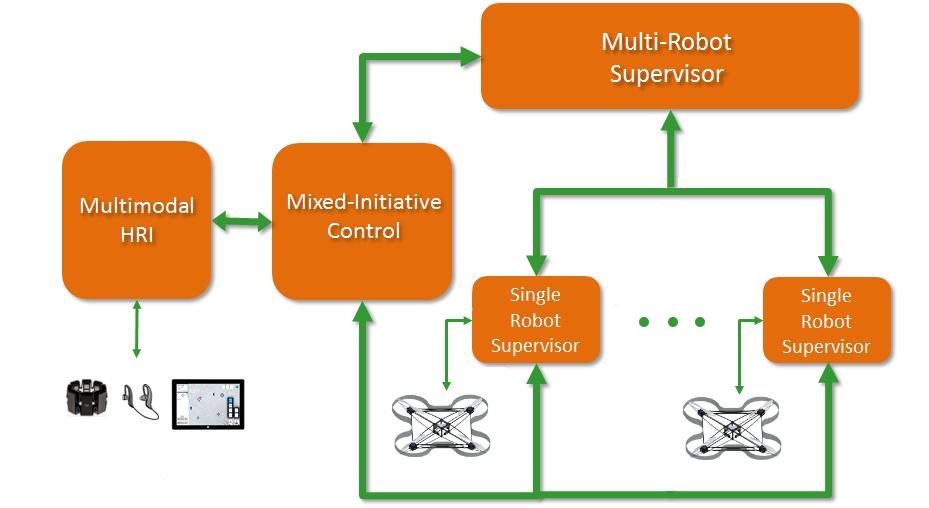}
		\par
		\caption[Complete Architecture]{The overall HRI architecture}
		\label{fig:completearch}
	\end{center}
\end{figure}

The operator should be capable of interacting with the system using different modalities (gestures, speech, tablet, etc.) at different levels of abstraction (task, activity, path, trajectory, motion, etc.). These continuous human interventions are to be suitably and reactively interpreted and integrated in the robotics control loops providing a natural and intuitive interaction.  
The HRI architecture designed to accomplish these requirements is illustrated in Fig.~\ref{fig:completearch} whose components are detailed below.

The Multimodal Interaction Module interprets the operator commands/intentions integrating inputs from multiple communication channels. For instance, either speech- or gesture-based commands may be used to stop a drone, while vocal commands in combination with deictic gestures can be used to specify navigational commands (e.g.  ``{\em go-there}'') to the co-located drones. Notice that commands can be given at different levels of abstraction, from task assignments  to direct teleoperation, combining different modalities. 
On the other hand, beyond line-of-sight control, the human can receive feedback from the drones via a tablet interface, the headphones, and the armband.

Once interpreted, the multimodal human commands are managed by the Mixed-Initiative Control module that interacts with the Single Robot Supervisor and Multirobot Supervisors mediating between the human and the robotic initiative. Indeed, commands can be provided to both single or multirobots, while vague instructions are to be completed and instantiated by the robots supervisory systems according to the operational context. Following a mixed-initiative planning approach \cite{Finzi05}, the human interventions are integrated into continuous planning and execution processes that reconfigure the robotic activities according to the human intentions and the operative state.
We refer the reader to \cite{BevacquaCFL15} for details. 
   
\begin{table}[h]
	\caption{List of primitive commands with modalities}
		\begin{center}
		\begin{tabular}{|c|c|}			
			\hline
			\textbf{Command} 		&  \textbf{Modalities} \\
			\hline
			\textit{Take-Off} 		& speech \\			
			\hline	
			\textit{Continue} 		& speech/gesture \\			
			\hline	
			\textit{Land} 			&  speech \\			
			\hline	
			\textit{Rotate \#o'clock}   &  speech/gesture and speech \\
			\hline	
			\textit{Selection \#Drone-Id}   &    speech/gesture and speech \\
			\hline	
			\textit{Faster}   &  speech/gesture \\
			\hline	
			\textit{Slower}   &  speech/gesture \\
			\hline			
			\textit{Rotate Clockwise} & speech/gesture \\
			\hline	
			\textit{Rotate Anti-clockwise} &  speech/gesture \\
			\hline					
			\textit{Brake} 			& speech/gesture\\			
			\hline	
			\textit{Go \#Direction}  &  speech/gesture/gesture and speech \\
			\hline	
			\textit{Search Expanding}   &  speech/gesture \\
			\hline	
			\textit{Search Parallel Track}   & speech/gesture \\
			\hline	
			\textit{Search Creeping Line}   &  speech/gesture \\
			\hline
			\textit{Switch}   &  gesture \\
			\hline									
		\end{tabular}
		\label{tab:voice_gesture_command}
		\end{center}
\end{table}

%
%

\subsection{Multimodal Interaction}
\begin{figure}[h]
	\begin{center}
		\includegraphics[scale=0.33]{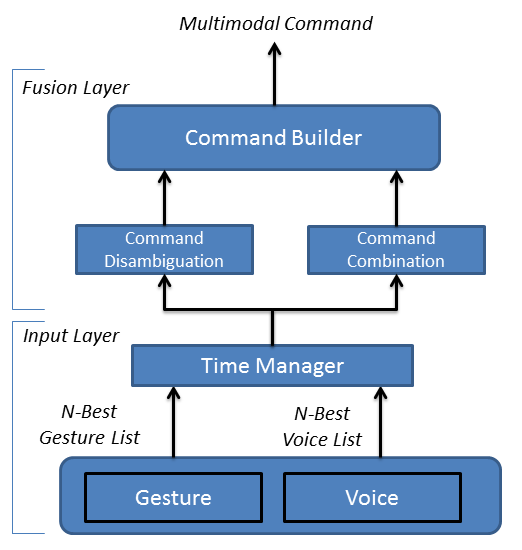}
		\par
		\caption[Multimodal Architecture]{Multimodal Interaction System}
		\label{fig:multimodal}
	\end{center}
\end{figure}

We employ a multimodal interaction framework that exploits a late fusion approach \cite{iros2013} where single modalities are separately classified and then combined (see Fig.~\ref{fig:multimodal}), this approach permits to introduce new modalities (e.g. tablet-based interaction) in an extensible and modular manner.

\begin{figure}[h]
	\begin{center}
		\includegraphics[scale=0.06]{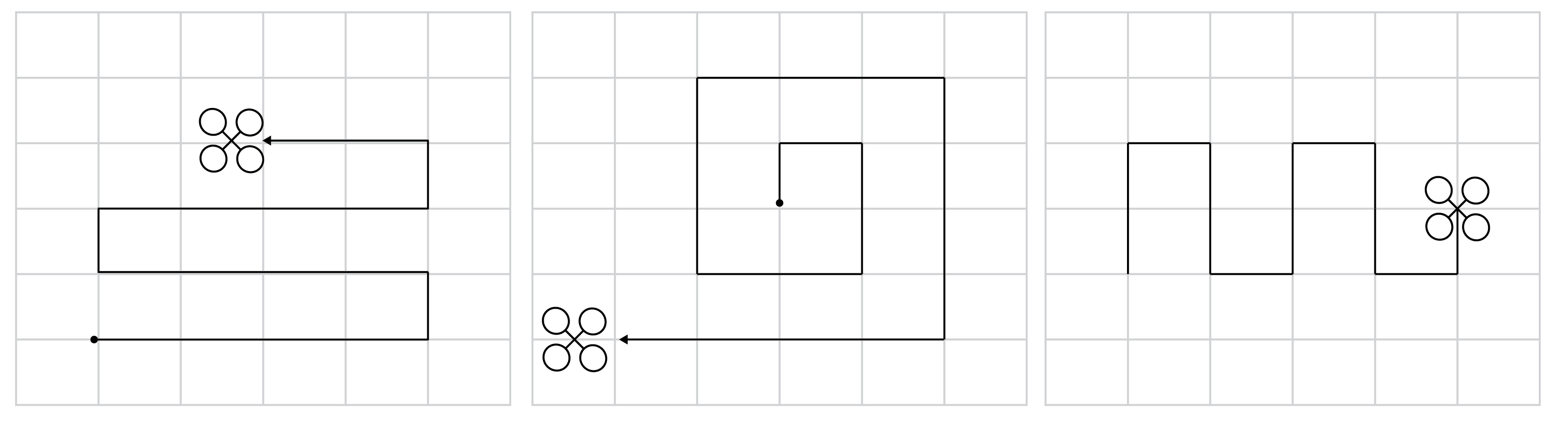}
		\par
		\caption[searching_patterns]{Parallel track (Left), Expanding (Center) and Creeping line (Right) search patterns}
		\label{fig:search_patterns}
	\end{center}
\end{figure}

In this work, we mainly focus on commands suitable for interacting with the set of co-located drones during navigation and search tasks. 
In particular, we are concerned with speech- and gestures-based communication with the drones suitable for the following purposes:
\begin{itemize}
\item
{\em Selection:} in order to select single or groups of robots involved in the action, both speech (e.g.  ``{\em all hawks take off}'',  ``{\em red hawks land}'') and gestures (e.g.  ``{\em you go down}'') can be used in combination. Specific names (e.g. ``{\em red hawk}'',  ``{\em blue hawk}'', etc.) can identify drones, while deictic gestures (pointing) can be used to select not only drones, but also targets (e.g.  ``{\em you go there}'').
\item
{\em Motion:} a set of commands are used for navigation. 
Motion directives can be coupled with voice directives. For example, a rotation  command (gesture) can be associated with the final orientation (voice) of the drone, while during a movement command (gesture) the operator can specify the exact distance to be covered (voice). When these values are not explicitly provided, default ones are assumed.
\item
{\em Search:} a set of commands are used to select the search pattern used to scan an area with a helicopter search \cite{NATO-88}, i.e. search-expanding, search-parallel-track and search-creeping-line (see Fig.~\ref{fig:search_patterns}). Those patterns can be  invoked either vocally or by means of specific gestures.
\item 
{\em Switch:} meta level commands allow the operator to change the interaction mode, e.g. from command-based to joystick-like interactive control and viceversa.
\end{itemize}

Table \ref{tab:voice_gesture_command} summarizes the overall set of primitive commands that we have introduced in our domain. These can be  invoked and flexibly combined in a multimodal manner using speech, gestures, or speech and gestures together.
In the following, we provide details about the adopted methods for speech/gesture recognition and fusion. 

\paragraph*{Speech recognition} we rely on Julius\cite{julius01}, a two-pass large vocabulary continuous speech recognition (LVCSR) engine. 
A suitable grammar has been defined to parse the commands of the users. A N-best list of possible interpretations is continuously provided in
output.   

\begin{figure}[h]
	\begin{center}
		\includegraphics[scale=0.35]{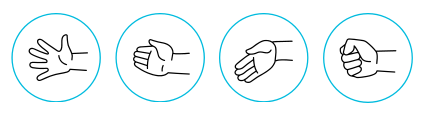}
		\par
		\caption[myo poses]{Different poses of the hand recognized by Thalmic Myo Armband.}
		\label{fig:myo_poses}
	\end{center}
\end{figure}

\begin{figure}[h]
	\begin{center}
		\includegraphics[scale=0.18]{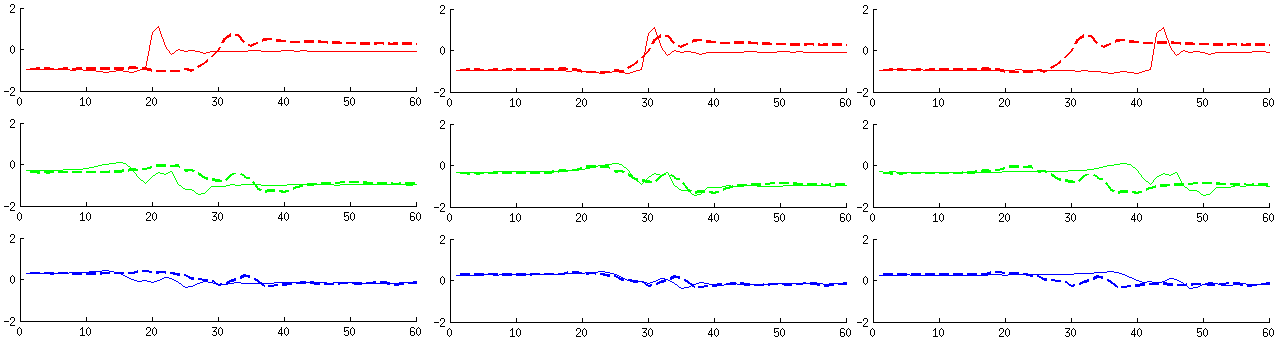}
		\par
		\caption[Sliding Windows GR]{Gesture matching at different time (start, middle, end) of the sliding window. Dashed and solid curves represent the recognized and executed gesture respectively.}
		\label{fig:sliding_window_gr}
	\end{center}
\end{figure}

\paragraph*{Gestures recognition}
The proposed gesture recognition system exploits the {\em Thalmic Myo Armband}. This device permits to detect and distinguish several poses of the hand (see Fig.~\ref{fig:myo_poses}) from the electrical activity of the muscles of the arm where the band is weared. In addition, 
the band is endowed with a $9$ $DOF$ IMU that permits motion capture. Therefore, both hand poses and movements
can be detected and used for robot control. 
In our framework, the position of the hand is used to enable/disable control modalities ({\em switch} commands), while
the movements of the hand are to be interpreted as gestures. Specifically,
in this work we introduce the following intuitive switching strategy: when the hand is closed (Fig. \ref{fig:myo_poses}, right) the command-based gesture interpretation is enabled, when the hand is open (Fig..~\ref{fig:myo_poses}, left) the joystick-like control is active.
As far as the hand poses are concerned, we directly rely on the built-in {\em Myo Armband} classifier. Instead, a robust gesture recognition system based on the armband acceleration measures requires an independent classification method. Specifically, our approach is inspired by the one of \cite{Wobbrock2007} and \cite{kratz2010}, which relies on geometric methods to classify. The main advantage of this method is that few examples are needed for training, while ensuring a robust user-dependent application. Since in our scenario we assume the presence of a trained operator (the busy genius) with a tailored recognition system, this approach is satisfactory. On the other hand, continuous recognition is not supported: the classifier needs the start and the end of the executed gestures. 
However, as already mentioned, we can exploit the hand pose detected by the {\em Myo Armband} to enable and disable the classifier, indeed we assume that the samples of a gesture are stored and classified only when the hand is closed ({\em switch} command).
Once the samples of an executed gesture are collected, the gesture classification process works as follows. 
Initially, two transformations are deployed to filter the acceleration signal generated by the input device. The first one removes noise 
from the acceleration samples. 
The second filter allows a uniform sampling independently of the execution speed of the gesture. 
For this purpose, analogously to \cite{kratz2010}, we linearly interpolate $m$ points in order to transform the input signal into a $m$-pla of samples $\langle \vec a_1,\dots, \vec a_m \rangle$ of fixed size with equal distance between them.  
After this phase, a third transformation makes the gesture invariant from the actual position of the arm of the operator and compensates the gravity force. Specifically, each sample is modified as: $\vec a^{'}_i $ = $R_S^W \vec a_i - \vec g$, with $R_S^W$ the rotation matrix from the sensor reference frame $S$ to the world frame $W$ and $\vec g$ the gravity acceleration vector. This way, each gesture $G$ can be associated with a uniform $m$-pla of samples. Given a training set $T_s$ that collects a set of $m$-ple for each gesture type, the gesture classification can be directly obtained form an Euclidean distance in the 3D space. Specifically, given the performed gesture $G$, for each gesture $G_i$ $\in$ $T_s$, we can define a score $s_i = || G - G_i ||$. The best scores for each gesture type defines the $N$-best list of matches for the executed gesture.
Since we need a reliable classifier, we improved the robustness of the \cite{kratz2010} classifier in two ways. First of all, for each executed gesture we generate not only one $m$-pla,  but a set of $m$-ple representing possible slides of the gesture samples within a time window (see Fig. \ref{fig:sliding_window_gr}). The best match is then used to rank the gesture with respect to the training set. 
In the second place, a secondary evaluation that takes into account the orientation of the executed gesture is performed when the difference between the last $2$-best values is below a suitable threshold. If this is the case, the gestures are ranked again by improving the values of the gestures whose average orientation is closer to the average orientation of the performed gesture. 

As for the dataset, we defined $14$ different types of gestures (with $10$ trials for each type) for both  navigation commands and search strategy requests. We evaluated the gesture recognizer by performing $30$ trials of each gesture (randomly generated to avoid the learning effect). In Table \ref{tab:gesture_classification_res}, we report the \textit{Precision}, \textit{Recall}, and \textit{Accuracy} for the gestures of the training set. 

\begin{table}[h]\footnotesize
	\caption{Gesture classification results}
		\begin{center}
		\begin{tabular}{|c||c||c|c||}
			\hline
			\textbf{Gesture} & \textbf{Precision} & \textbf{Recall} & \textbf{Accuracy}\\
			\hline
			\textit{Brake} & 93.3\% & 96.5\% & 94.9\% \\
			\hline			
			\textit{Go Ahead} & 96.6\% & 90.6\% & 93.4\% \\
			\hline
			\textit{Go Backward} & 100\% & 100\% & 100\% \\
			\hline
			\textit{Go Down} & 100\% & 88.3\% & 93.5\% \\
			\hline
			\textit{Go Left} & 73.3\% & 100\% & 86.2\% \\
			\hline
			\textit{Go Right} & 96.6\% & 85.3\% & 90.4\% \\
			\hline
			\textit{Go Up} & 100\% & 96.7\% & 98.3\% \\
			\hline
			\textit{Rotate Anti-Clockwise} & 100\% & 96.7\% & 98.3\% \\
			\hline
			\textit{Rotate Clockwise} & 83.3\% & 100\% & 91.38\% \\
			\hline
			\textit{Search Creeping Line} & 80\% & 82.7\% & 82.4\% \\
			\hline
			\textit{Search Expanding} & 96.6\% & 85.3\% & 90.4\% \\
			\hline
			\textit{Search Parallel Track} & 96.6\% & 90.6\% & 93.4\% \\
			\hline
			\textit{Faster} & 96.6\% & 82.9\% & 88.9\% \\
			\hline
			\textit{Slower} & 66.6\% & 90.9\% & 79.9\% \\
			\hline
			\hline
			\textit{Average} & 91.4\% & 91.9 \% & 91.7 \% \\
			\hline
		\end{tabular}
		\label{tab:gesture_classification_res}
	\end{center}
\end{table}
\begin{table}[h]
	\caption{Multimodal classification results}
		\begin{center}
		\begin{tabular}{|c||c||c|c||}
			\hline
			\textbf{} & \textbf{Precision} & \textbf{Recall} & \textbf{Accuracy}\\
			\hline
			\textit{Average}:  & 96.95\% & 96.4\% & 96.3\%\\
			\hline

		\end{tabular}
		\label{tab:gesture_classification_res}
	\end{center}
\end{table}

\paragraph*{Multimodal Fusion}
The {\em fusion module} combines the results of vocal and/or gesture recognition providing the command interpretation. We deploy a late fusion approach that exploits the confidence values generated by the separated (speech and gestures) classifiers. In order to integrate the classifiers outcomes, the two channels are to be first synchronized. We assume that the first channel that becomes active (speech or gesture) starts a time interval (about $1$ sec in our setting) during which any other activity can be considered as synchronized. This way, a vocal command provided during the execution of a gesture, or immediately, after can be fused (or a gesture after a vocal command).  When this is the case, contextual rules are used to \textit{disambiguate} the conflicting commands or to \textit{combine} vocal and gesture inputs using the information contained in both the channels.  In the first case, or when an explicit rule is not available to disambiguate, the $N$-best values provided by the speech and gesture recognizers are compared and the interpretation with the maximum value is selected. 
In the second case, if the classification results are compatible, these are combined according to simple rules (e.g. navigation gesture towards a certain direction combined with a vocal indication of a distance is interpreted as \textit{Go to \#distance}). In Table \ref{tab:gesture_classification_res}, the average classification results - collected by rerunning the classifier evaluation considering the fusion of speech and gestures - show the improvement in robustness due to multimodal disambiguation. Overall, the reliability and the latency of the multimodal interaction system seems compatible with a natural and effective interaction.

\subsection{Mixed-Initiative Interaction}
The operator is allowed to interact with the drones at any time at different levels of abstraction with different interaction metaphores.  
In this work, we focus on navigation and search activities, hence our main concern is on path/trajectory level interaction (for task level interaction see \cite{BevacquaCFL15}).  In this context, we introduced two main interaction metaphores: {\em command-based} and {\em joystick-based}. In the first case the robots are considered as agents to be coordinated by the operator multimodal commands, in the second case the operator can directly teleoperate a drone using his/her open hand as a virtual joystick. Here, the hand position is used to switch between these two control modes: {\em hand-closed} for gestural commands, {\em hand-open} for trajectory adjustments and teleoperation.
For instance, the human may ask a drone the execution of a task (e.g. ``blue hawks go there''), and once the execution starts, use the {\em hand-open} mode to manually correct the trajectory. A complete teleoperated control can be obtained once the current command execution has been stopped by a {\em brake} command.   

\paragraph*{Path and Trajectory level interaction}
The navigation commands introduced so far are associated with drone movements to be suitably planned and executed. We deploy a $RRT^*$ algorithm \cite{Karaman.Frazzoli:RSS10} for path planning, while trajectory planning is based on a 4-$th$ order spline concatenation preserving continuous acceleration. Following the approach in \cite{iros14}, during the execution, the human is allowed to on-line adjust the robot planned trajectory without provoking replanning ({\em open-hand} mode). Specifically, the planned robot position $a(t)$ can be deviated into a mixed trajectory $m(t) = a (t) + h (t)$, by the $h(t)$ human contribution defined as follows
\begin{center}
			$h$(t) = $\begin{cases}
			h(t-1)+human(t) $ if $ mixed=ON\\
			h(t-1)+\Lambda(t) $  $ otherwise
		\end{cases}$
\end{center}
That is, when the mixed-initiative is active the control reference $human(t)$ generated  by the operator -via gestures or voice- increases the current displacement $h(t)$; otherwise, when the human intervention is released, the deviated trajectory is smoothly driven back towards the planned one by a linear function $\Lambda(t)$. 
%
%
On the other hand, similarly to \cite{iros14}, if the human deviation goes outside a context-specific workspace, a replanning process starts and the autonomous system generates another path and another trajectory to reach the next waypoint. A vibro-tactile feedback on the armband provides the operator with the perception of the robot deviation with respect to the planned trajectory. 


\section{CASE STUDY}
\begin{figure}[h]
	\begin{center}
		\includegraphics[scale=0.165]{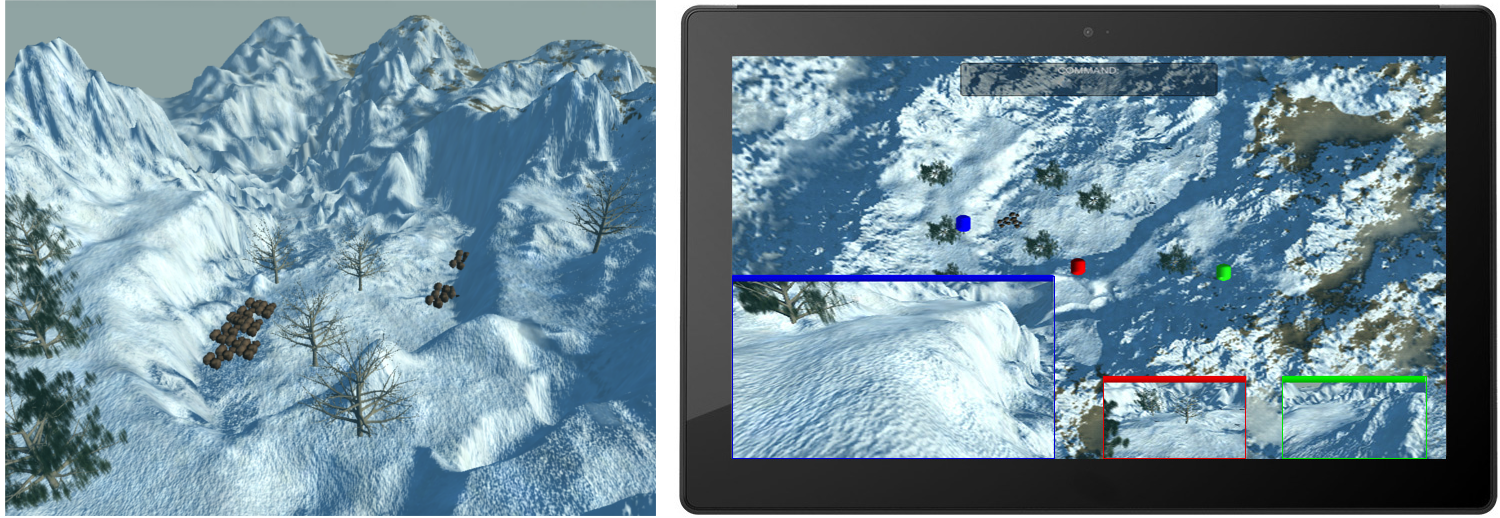}
		\par
		\caption[user_interface]{Simulated environment (left), tablet interface (right).}
		\label{fig:user_interface}
	\end{center}
\end{figure}

The effectiveness of the multimodal interaction framework proposed in this work has been tested in a simulated environment of an alpine scenario (see Fig.~\ref{fig:user_interface}, left). We used \textit{Unity 3D} to simulate a set of drones 
equipped with an onboard camera. A tablet user interface (see Fig.~ \ref{fig:user_interface}, right) allows the operator to monitor the robots position on a map, while receiving video streams for the cameras of the drones on multiple windows; the one associated with the selected drone has a bigger size. 
%
In this scenario, a set of victims is randomly positioned within the environment. The mission goal is to find a maximum number of missed persons within a time deadline. When a missed person appears in the field of view of a drone camera, the operator can use the tablet interface to mark his/her position.  

\begin{figure}[h]
	\begin{center}
		\includegraphics[scale=0.05]{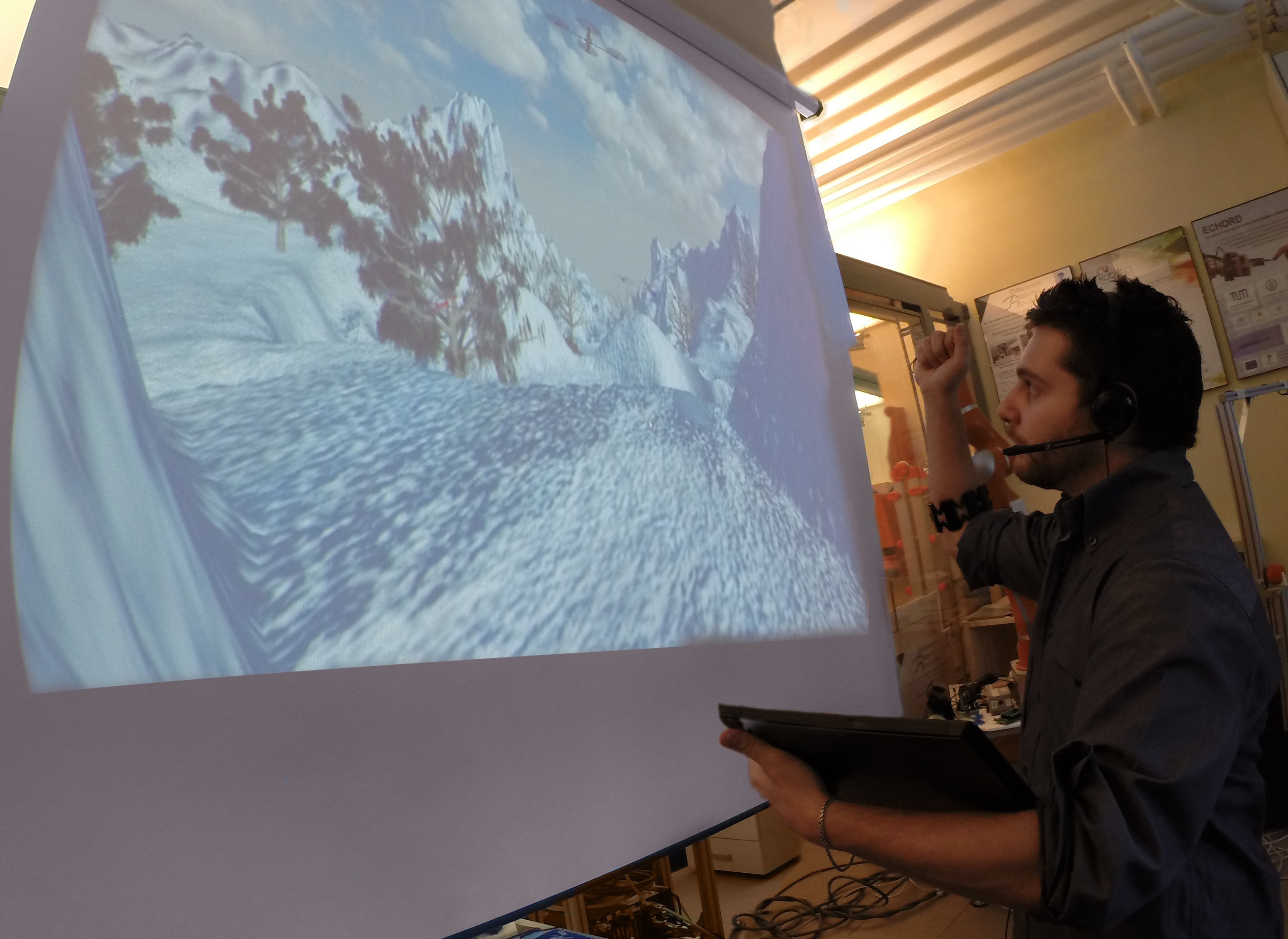}				
		\par
		\caption[alpine_scenario]{Simulated alpine scenario testbed. A video  demostration can be found at http://wpage.unina.it/jonathan.cacace/Media/roman2016-i.mp4}
		\label{fig:alpine_scenario}
	\end{center}
\end{figure}

\paragraph{Experimental set-up}
As an pilot study, we designed an experimental set-up with the aim of reproducing a similar setting in the real world. Preliminary experience with our real drones suggested to start form an initial configuration with $2$ or $3$ UAVs per operator. We focus on a winter scenario, where the rescuers operate under time pressure in a restricted area, therefore we defined a $120 \times 120$ $m$ with the goal of searching for $6$ missing persons in a maximum time of $6$ minutes. The target user is a trained operator, hence we involved a group of $5$ expert users ($4$ males, $1$ female). Each subject was asked to perform $3$ runs of a mission. 
For each trial we collected the following data: 
\begin{itemize}
	\item \textit{detected persons}: number of discovered victims;
		\item \textit{time to detect}: mean time needed to find a victim;
	\item \textit{selection time}: time spent while monitoring and controlling a drone;
	\item \textit{operative mode}: time spent per drone for each operative mode (autonomous, mixed-initiative, teleoperation);
	\item \textit{interaction type}: modality used to invoke commands (voice, gesture, voice and gestures, etc.).
\end{itemize}

\paragraph{Results}
In Table \ref{tab:total_results}, we report the victims found (mean values) using $2$ and $3$ drones along with the mean, maximum, and minimum time (sec) needed to find a victim. We further detail the user performance in Table \ref{tab:detailed_results} where we illustrate the success rate in finding $n$ (of $6$) victims along the mean time (sec) needed. We can observe that the overhead of monitoring and controlling $3$ drones seems here compensated by better performances in search, indeed the $3$ drones search outperforms the $2$ drones one ($5.6$ vs. $3.9$ victims found, with two-tailed p $<.0001$), while the mean time needed to find the $n$-th victim (e.g. $96,6$ vs. $249,6$ sec. to find the first $3$ victims, two-tailed p $<.0001$) is also considerably reduced. Table \ref{tab:two_drones} shows how the operators balance the time (sec) dedicated to the $3$ drones during the tests. Here, we consider: maximum ({\em max}), minimum ({\em min}), middle ({\em mid}) time dedicated to a drone, and the time spent monitoring all the drones ({\em all}). The averages of these values and the associated selection percentage, shows a satisfactory balance. Interestingly, the time dedicated to {\em all} the drones is not negligible, indeed all drones are selected during parallel operations (e.g. ``all hawks take off'') or to inspect all the cameras at the same time during a high-level scan. 

\begin{table}[h] 
            \caption{Mission results: victims found and time to find a victim}
        \begin{center} 
        {\footnotesize
        \begin{tabular}{|c || c | c | c | c | c | c | c | c |}  
        \hline
         & \multicolumn{4}{|c|}{{\bf Victims}} & \multicolumn{4}{|c|}{{\bf Time}}  \\ \cline{2-9}
         & avg & min & max & std & avg & min & max & std \\ \hline
         {\bf 2 Drn} & 3.9 & 3 & 5 & 0.65 & 217.4 & 60 & 350 & 18.3 \\ \hline
         {\bf 3 Drn} & 5.6 & 4 & 6 & 0.61 & 157.6 & 45 & 350 & 17.1  \\ \hline  
        \end{tabular}
        }
        \label{tab:total_results}
        \end{center}
\end{table}




\begin{table}[h] \footnotesize
            \caption{Mission details: time to find $n$ victims}
        \begin{center}
        \begin{tabular}{|c c||c|c|c|c|c|c|}            
            \hline
            \multicolumn{2}{|l||}{{\bf 2 Drn}}   & 1 & 2 & 3 & 4 & 5 & 6  \\
            \hline
           \multicolumn{2}{|l||}{{\bf Succ.} \%}  & 100 & 100 & 100 & 80 & 20  & 0 \\ \hline
             \multirow{2}{*}{{\bf Time}} & \multicolumn{1}{ l|| }{avg}  & 85.8 & 122.7  & 249.6 & 296   & 333.3 & --   \\ \cline{3-8}
                   & \multicolumn{1}{ l|| }{std} & 15.7 & 22.8  & 24 & 13.8  & 15.2 & --  
                              \\
            \hline \hline
             \multicolumn{2}{|l||}{{\bf 3 Drn}}  & 1 & 2 & 3 & 4 & 5 & 6  \\
            \hline
           \multicolumn{2}{|l||}{{\bf Succ. } \%}  & 100 & 100 & 100 & 100 & 93.3 & 46,6  \\ \hline
           \multirow{2}{*}{{\bf Time} } & \multicolumn{1}{ l||}{avg}     & 59 & 64.3 & 96.6 & 166.9 & 235.5 & 324.2   
\\  \cline{3-8}            
           & \multicolumn{1}{ l|| }{std}   & 6.98 & 6.13 & 12.3 & 25.2 & 32.5 & 19.8   
\\             
            \hline                
        \end{tabular}
        \label{tab:detailed_results}
        \end{center}
\end{table}

\begin{figure}[h]
	\begin{center}
		\includegraphics[scale=0.3]{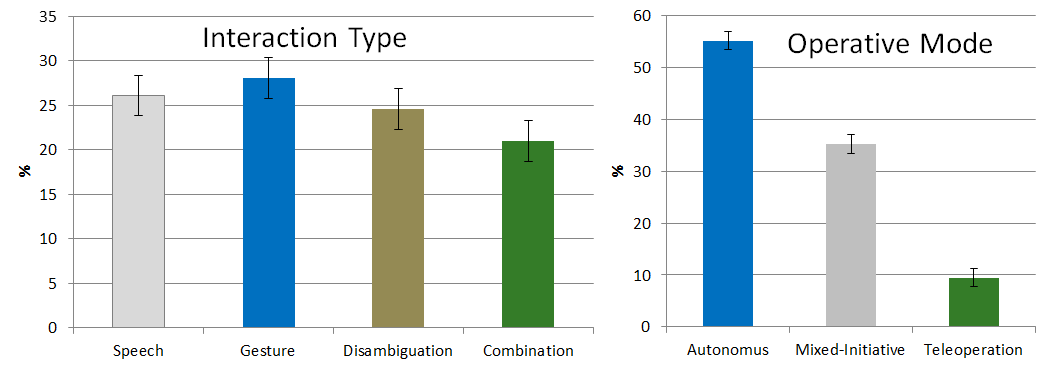}
		\par
		\caption[command_source]{Interaction type (left) and operative modes (right)}
		\label{fig:command_source}
	\end{center}
\end{figure}

We can also analyze whether the multimodal interaction is actually exploited. 
For this purpose, Figure \ref{fig:command_source} (left) illustrates the distribution of the commands for each modality. Interestingly, we can observe that the gesture-based commands are frequently used to orchestrate the operations of the drones during the search, both as single gestures, or in combination with vocal instructions (used either to complete or to reinforce a command), while purely vocal interactions are less frequent.   
%
%
\begin{table}[h]
	\caption{Drone Selection: time dedicated to the drones}
		\begin{center}
		\begin{tabular}{| c c ||c|c|c|c|}			
			\hline
			 & & Max & Mid & Min & All \\
			\hline
			\multirow{2}{*}{{\bf Time} } & avg   & 108 & 92.16 & 80.64 &  75.6 \\ \cline{3-6}
			 & std  & 17.9 & 16.3 & 17.6 & 31.6 \\
			\hline
			\multicolumn{2}{|l||}{{\bf Select. } \%} & 30 & 25.6 & 22.4 & 22 \\
			\hline
 
			\end{tabular}
		\label{tab:two_drones}
		\end{center}
\end{table}
Finally, in Figure \ref{fig:command_source} (right) we compare the precentage of time spent by the drones in the operative modes ({\em autonomous}, {\em teleoperated}, {\em mixed-initiative}). Here, we can observe that during the tests ($3$ drone cases) each drone mainly operates in the autonomous mode, with a minor percentage of time spent in interactive adjustments (joystick-based interaction), while the direct teleoperation is a rare control modality. The dominance of the autonomous mode allows the user to sporadically interact with the drones as required in the SHERPA scenario.
%
Overall, this preliminary evaluation suggests that the proposed multimodal interaction framework is effective in the $3$ drone configuration, indeed, the user monitoring and control effort during search mission seems well balanced among the drones, while both the multimodal interaction and the autonomous control mode seem exploited as expected. 

\section{CONCLUSIONS}
We presented a multimodal human-robot interaction framework suitable for a rescuer that cooperates with a set of drones while searching for missing persons in an alpine scenario. We illustrated the overall interaction framework designed for this challenging domain discussing its features in a simulated case study. The initial evaluation shows that the proposed interaction system allows the human to flexibly interact with the drones in a reliable and effective manner exploiting different modalities. These preliminary results encourage us towards more extensive evaluation tests in similar real-world scenarios.




\section*{ACKNOWLEDGMENT}
The research leading to these results has been supported by the FP7-ICT-600958 SHERPA project.



\bibliographystyle{IEEEtran}
\bibliography{biblio-mixed}

\begin{thebibliography}{10}
\providecommand{\url}[1]{#1}
\csname url@samestyle\endcsname
\providecommand{\newblock}{\relax}
\providecommand{\bibinfo}[2]{#2}
\providecommand{\BIBentrySTDinterwordspacing}{\spaceskip=0pt\relax}
\providecommand{\BIBentryALTinterwordstretchfactor}{4}
\providecommand{\BIBentryALTinterwordspacing}{\spaceskip=\fontdimen2\font plus
\BIBentryALTinterwordstretchfactor\fontdimen3\font minus
  \fontdimen4\font\relax}
\providecommand{\BIBforeignlanguage}[2]{{%
\expandafter\ifx\csname l@#1\endcsname\relax
\typeout{** WARNING: IEEEtran.bst: No hyphenation pattern has been}%
\typeout{** loaded for the language `#1'. Using the pattern for}%
\typeout{** the default language instead.}%
\else
\language=\csname l@#1\endcsname
\fi
#2}}
\providecommand{\BIBdecl}{\relax}
\BIBdecl

\bibitem{Sherpa2}
L.~Marconi, C.~Melchiorri, M.~Beetz, D.~Pangercic, R.~Siegwart, S.~Leutenegger,
  R.~Carloni, S.~Stramigioli, H.~Bruyninckx, P.~Doherty, A.~Kleiner,
  V.~Lippiello, A.~Finzi, B.~Siciliano, A.~Sala, and N.~Tomatis, ``The sherpa
  project: Smart collaboration between humans and ground-aerial robots for
  improving rescuing activities in alpine environments,'' in \emph{Proc. of
  SSRR-2012}, pp. 1--4.

\bibitem{Cummings2007}
M.~Cummings, S.~Bruni, S.~Mercier, and P.~J. Mitchell, ``Automation
  architecture for single operator, multiple uav command and control,''
  \emph{Intern. Command and Control Journal}, vol.~1, no.~2, pp. 1--24, 2007.

\bibitem{Bitton2008}
E.~Bitton and K.~Goldberg, ``Hydra: A framework and algorithms for
  mixed-initiative uav-assisted search and rescue,'' in \emph{Proc. of CASE
  2008}, 2008, pp. 61--66.

\bibitem{MazaCMPO10}
I.~Maza, F.~Caballero, R.~Molina, N.~Pena, and A.~Ollero, ``Multimodal
  interface technologies for uav ground control stations.'' \emph{Journal of
  Intelligent and Robotic Systems}, vol.~57, no. 1-4, pp. 371--391, 2010.

\bibitem{PfeilKL13}
K.~Pfeil, S.~L. Koh, and J.~LaViola, ``Exploring 3d gesture metaphors for
  interaction with unmanned aerial vehicles.'' in \emph{Proc. of IUI 2013},
  2013, pp. 257--266.

\bibitem{szafir2014communication}
D.~Szafir, B.~Mutlu, and T.~Fong, ``Communication of intent in assistive free
  flyers,'' in \emph{Proc. of HRI 2014}, 2014, pp. 358--365.

\bibitem{NagGiuGamDic14:iros}
J.~Nagi, A.~Giusti, L.~Gambardella, and G.~A. {Di Caro}, ``Human-swarm
  interaction using spatial gestures,'' in \emph{Proc. of IROS 2014}, 2014, pp.
  3834--3841.

\bibitem{MonajjemiWVM13}
V.~M. Monajjemi, J.~Wawerla, R.~T. Vaughan, and G.~Mori, ``Hri in the sky:
  Creating and commanding teams of uavs with a vision-mediated gestural
  interface.'' in \emph{Proc. of IROS 2013}, 2013, pp. 617--623.

\bibitem{iros14}
J.~Cacace, A.~Finzi, and V.~Lippiello, ``A mixed-initiative control system for
  an aerial service vehicle supported by force feedback,'' in \emph{Proc. of
  IROS 2014}, 2014, pp. 1230--1235.

\bibitem{BevacquaCFL15}
G.~Bevacqua, J.~Cacace, A.~Finzi, and V.~Lippiello, ``Mixed-initiative planning
  and execution for multiple drones in search and rescue missions,'' in
  \emph{Proc. of ICAPS 2015}, 2015, pp. 315--323.

\bibitem{Finzi05}
A.~Finzi and A.~Orlandini, ``Human-robot interaction through mixed-initiative
  planning for rescue and search rovers,'' in \emph{Proc. of AI*IA 2005}, 2005,
  pp. 483--494.

\bibitem{iros2013}
S.~Rossi, E.~Leone, M.~Fiore, A.~Finzi, and F.~Cutugno, ``An extensible
  architecture for robust multimodal human-robot communication,'' in
  \emph{Proc. of IROS 2013}, 2013, pp. 2208--2213.

\bibitem{NATO-88}
NATO, \emph{ATP-10 (C). Manual on Search and Rescue. Annex H of Chapter 6},
  1988.

\bibitem{julius01}
A.~Lee, T.~Kawahara, and K.~Shikano, ``Julius--an open source real-time large
  vocabulary recognition engine,'' in \emph{Proc. of EUROSPEECH-2001}, 2001.

\bibitem{Wobbrock2007}
A.~W. J.O.~Wobbrock and Y.~Li, ``Gestures without libraries, toolkits or
  training: A \$1 recognizer for user interface prototypes,'' in \emph{Proc. of
  the UIST}, 2007, pp. 159--168.

\bibitem{kratz2010}
S.~Kratz and M.~Rohs, ``A \$3 gesture recognizer: simple gesture recognition
  for devices equipped with 3d acceleration sensors,'' in \emph{Proc. of
  IUI-10}, 2010, pp. 341--344.

\bibitem{Karaman.Frazzoli:RSS10}
S.~Karaman and E.~Frazzoli, ``Incremental sampling-based algorithms for optimal
  motion planning,'' in \emph{Robotics: Science and Systems (RSS)}, 2010.

\end{thebibliography}

\end{document}